# Transformer-Based Classification Outcome Prediction for Multimodal Stroke Treatment

Danqing Ma*
Computer Science
Stevens Institute of Technology
Hoboken NJ, U.S
*madanqing376@gmail.com

Meng Wang
Department of Computer Science
University of Sofia
Palo Alto, CA, U.S

Ao Xiang
School of Computer Science & Engineering
University of Electronic Science and Technology of China
Chengdu, Sichuan, China

Zongqing Qi
Computer Science, Stevens Institute of Technology,
Hoboken NJ, U.S

Qin Yang
School of Integrated Circuit Science and Engineering
University of Electronic Science and Technology of China
Chengdu, Sichuan, China

*Abstract*—This study proposes a multi-modal fusion framework Multitrans based on the Transformer architecture and self-attention mechanism. This architecture combines the study of non-contrast computed tomography (NCCT) images and discharge diagnosis reports of patients undergoing stroke treatment, using a variety of methods based on Transformer architecture approach to predicting functional outcomes of stroke treatment. The results show that the performance of single-modal text classification is significantly better than single-modal image classification, but the effect of multi-modal combination is better than any single modality. Although the Transformer model only performs worse on imaging data, when combined with clinical meta-diagnostic information, both can learn better complementary information and make good contributions to accurately predicting stroke treatment effects.

## I. INTRODUCTION

Stroke refers to a range of disorders caused by occlusion or haemorrhage of blood vessels supplying the brain. Acute ischaemic stroke is the most common type of stroke and is one of the leading causes of disability and death worldwide [1]. This is a disease caused by the formation of blood clots when blood flow to the brain is interrupted. If the blockage is not resolved, the extent of dead tissue increases, resulting in loss of brain cells from irreversible damage [2]. Because of the high mortality and disability rates associated with stroke, it is crucial to emphasise timely and appropriate treatment. Early detection and prevention of stroke is needed. Both manifestations of stroke, ischemic and hemorrhagic, are quite dangerous and require urgent medical attention. Prompt recognition of warning symptoms so that professional medical help can be sought within optimal treatment times is the primary goal. Ischaemic stroke cases were previously resolved through methods such as thrombectomy, but this is dependent on the severity of the stroke. It is important to note that the risk of brain haemorrhage and death still exists later in life [3] Therefore, determining whether newly admitted intermediate-grade patients can receive good treatment from mechanical thrombectomy is an important step to reduce the risk of complications and improve the quality of life of patients. Unfortunately, however. Although this is important, most past studies have ignored prognostic indicators in favour of treating the condition after it has presented. With the rise of artificial intelligence, the medical testing industry has ushered in a new era of intelligence [4]. Several machine learning algorithms are used to perform stroke detection, such as logistic regression [5], xgboost [6], support Vector Machines [7], while on the other hand, a number of image-based algorithms for stroke detection have been widely cited, including CNN [8], transformer [9] and other architectures.

These image-based algorithms have been applied to a variety of imaging modalities including Magnetic Resonance Imaging (MRI), Positron Emission Tomography (PET), non-contrast computed tomography (NCCT). Among the medical image detection currently studied, only a few studies have been synthesized for multimodality [10]. And there is a lack of using Transformer for multimodal data, especially images and text, for and stroke outcome prediction. By using deep learning and biosignals of various modalities. This method proposes a multimodal hybrid model based on a large model using diagnostic information provided by the hospital at the time of discharge and image information at the time of admission. To predict the functional outcome of stroke treatment based on the modified Rankin Scale (mRS). This model is able to efficiently combine NCCT images and diagnostic information. Our results on multimodal interaction are a valuable contribution to the advancement of multimodal deep learning methods in the field of stroke prediction.

## II. RELATED WORK

Machine learning has become an important part of credit assessment [11] business forecast [12], the mainstream algorithm for data analysis in many fields, and the same is true in medicine. Various aspects of algorithms for stroke detection have been fully discussed, for example Y Wu et al specifically focus on stroke prediction in elderly Chinese, demonstrating the effectiveness of machine learning methods in this population. [13], Regarding the imbalance of the dataset, and they proposed a unified embedding framework for stroke datasets to address the huge challenges posed by resolving imbalanced datasets. In addition to traditional structured data sets, in deep learning such as visual effects, in addition to the traditional CNN deep model, some architectures based on transformer architecture have gradually become including autonomous driving [14], Medical image recognition [15] and other research hotspots in many fields. A typical example, ViT [16], is a landmark vision-based pure transformer model. But this lacks modeling of local features and comes at the cost of huge training datasets that are not always available in specialized domains. TransUNet [17] is the first medical image segmentation framework. However, it only uses single-scale features for context modeling and ignores multi-scale feature dependencies. The latter often plays a key role in ischemic stroke segmentation because the lesion size changes drastically and cannot produce satisfactory results. Therefore, later, some improved algorithms such as EG-TransUNet [18], MLiRA-Net [19] and other new transformer algorithms have been proposed for improvement. The transformer algorithm is particularly suitable for multimodal tasks because of its unique architecture. In recent years, some stroke detection algorithms based on multimodal frameworks have been gradually investigated, e.g., Yoon, C et al. developed a deep learning-based segmentation model using multimodal UNet (MM-UNet) to predict the region of interest (ROI) in magnetic resonance (MR) images [20]. Samak, et al. invented a multi-modal architecture that uses a transformer architecture to inspect NCCT images and fuse them with clinical information [21]. Balázs Borsos et al. proposed a multimodal approach for predicting functional status of acute ischaemic stroke patients after hospital discharge based on tabular data and CT perfusion imaging [22]. Pustina et al. developed a simple multimodal architecture based on the transformer architecture to evaluate the sequelae of stroke [23]. The limitation of their study is that they still built some explicit frameworks for estimation and evaluation, rather than relying entirely on implicit feature interactions.

### III. METHODOLOGY

This paper proposes a multimodal detection architecture Multitrans. An overview is shown in Fig. 1, and this framework builds on previous work [24]. It consists of a two Transformer encoder architecture and a self-attention mechanism module for handling modal fusion. The data from all modalities are first represented as sequence data and positional tokens and learnable classification tokens (CLS) are added, after which they are fed into the transformer. A series of transformer blocks are used in the transformer encoder, each of which consists of a normalisation layer, a normalisation layer, and a multilayer perceptron (MLP) header followed by multi-head self-attention (MHSA). The MLP header is then applied to the classification labels to extract representations from images and text. Finally the different representations are spliced together and fed into the multimodal fusion module for further deep fusion, after which the final result is output after entering the MLP classifier.

The framework utilises BERT [25] and ViT as base models, as illustrated in Figure 2. A pre-trained BERT model is employed to process textual data. BERT is a natural language processing model that combines word embeddings, segment embeddings, and positional embeddings to create comprehensive embedding representations. Word embeddings map each word or subword to a vector in a high-dimensional space. Segment embeddings allow BERT to distinguish and process individual texts or text pairs. Positional embeddings play a crucial role in the Transformer architecture by providing sequential information that enables the model to understand the position of words in a sentence. Furthermore, BERT's CLS markup provides an efficient way to diagnose text for stroke. The Comprehensibility and Logical Structure characteristic requires that information be presented in simple form: the CLS is placed at the beginning of the input sequence, and its final hidden state after processing serves as a representation of the entire sequence, effectively representing the diagnostic state at the sentence level. For the image diagnosis part, we employ another classical model based on the Transformer architecture, ViT. ViT model is inspired by the BERT model, which transforms a vision task into a sequence modelling problem. It divides the input image into small patches and flattens them into a sequence, which is then processed by the Transformer model. We similarly adopt the same strategy for the images, using the CLS tags in the images to represent the meaning of the images. After that, the representations of the two modalities are spliced before going into feature fusion. The same strategy is adopted for the images, using the CLS tags to represent their meaning. The representations of the two modalities are then combined before feature fusion.

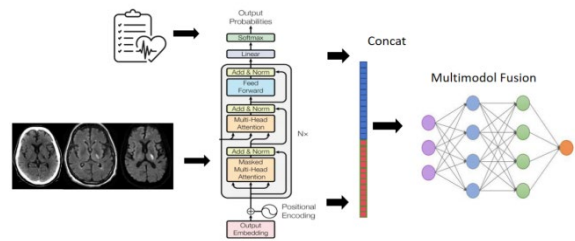

Fig. 1 Multimodal detection architecture of Multitrans.

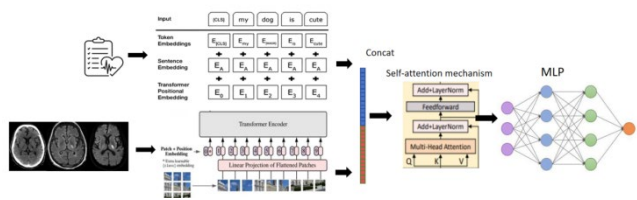

Fig. 2 The framework utilises BERT and ViT as base models.

The fusion feature module is composed of a self-attention mechanism module and a fully connected neural network. The self-attention layer computes the attentional weights of each element in the input over all other elements. Specifically, the self-attention layer takes the input vectors and generates three different sets of vectors by three different linear transformations: a Query vector, a Key vector, and a Value vector. For each element of the input the corresponding Q, K, and V vectors are generated. The self-attention layer computes the dot product of each query vector with all key vectors as the attention score. This means that the model will consider each part of the text and image information and compute their interactions, learning how to adapt its internal representation based on the global information of the text and image. The output of the self-attention mechanism is a vector with the same dimensions as the input. After which the output of the self-attention layer is fed directly into a linear layer for further processing, and ultimately passed to the Softmax layer for classification.

We collected the relevant diagnostic results and data from 128 patients of Capital Medical University for acute ischaemic stroke caused by intracranial artery occlusion. These data were self-supplied by the patients at a later stage after the questionnaire survey and are not official data. Forty-two of these patients received intra-arterial treatment and 86 patients received usual care. We segmented and trained all data using a 7:2:1 ratio. Outcome was judged by Modified Rankin Scale score at 90 days; this categorical scale measures functional outcome with scores ranging from 0 (no symptoms) to 6 (death). We dichotomised the mRS scores into one of the original 7 grades, where mRS ≤ 2 indicates a good outcome, and mRS > 2 indicates a poor outcome.

## IV. RESULT

In addition to performing performance tests on the benchmark models, we evaluated our Multitrans framework by performing ablation experiments on its encoder part using different Transformer architectures. The image part of the model uses the DeiT [26] and Swin Transformer [27] architectures. DeiT utilises ImageNet pre-trained DeiT models and knowledge distillation to both love roses migration learning, Multitrans -SwinT consists of two Swin transformer blocks and MHSA, where a pyramid-structured downsampling and sliding-window strategy is used to capture image details. The textual part uses RoBERTa [28] and ALBERT [29], which are become Multitrans -RoBERTaT and Multitrans -ALBERTT. Where RoBERTa is a more powerful performance with the training model, which employs the use of a larger dataset and a longer period of time for pre-training, as well as dynamically adjusting the MASK mechanism, among other things. ALBERT, on the other hand, is a more lightweight BERT-based modification that employs a parameter sharing mechanism to reduce the number of parameters and uses and introduces a sentence order prediction mechanism to reduce the model. We evaluate the classification performance of the model using three commonly used metrics: accuracy, F1 score and area under the ROC curve (AUC). Accuracy indicates the proportion of correctly predicted results to the total number of samples. F1 score is the reconciled mean of the precision rates Precision and Recall, and the AUC value indicates the area under the ROC curve, a graphical tool used to assess the performance of dichotomous classification models, with the horizontal axis being the False Positive Rate, and the vertical axis being the True Positive Rate. The results of the performance are shown in Table I.

TABLE I MODEL RESULT

| Method | Image-modal | Text-Modal | ACC | F1-score | AUC |
|---|---|---|---|---|---|
| Unimodal | None | BERT | 0.87 | 0.74 | 0.82 |
| Unimodal | None | RoBERTa | 0.85 | 0.72 | 0.82 |
| Unimodal | None | ALBERT | 0.87 | 0.76 | 0.87 |
| Unimodal | ViT | None | 0.56 | 0.42 | 0.65 |
| Unimodal | DeiT | None | 0.56 | 0.42 | 0.64 |
| Unimodal | SwinT | None | 0.56 | 0.42 | 0.66 |
| Multimodal | ViT | BERT | 0.89 | 0.75 | 0.84 |
| Multimodal | SwinT | BERT | 0.89 | 0.75 | 0.84 |
| Multimodal | SwinT | RoBERTa | 0.90 | 0.77 | 0.85 |

The experimental results demonstrate several important findings. Firstly, unimodal text categorisation shows similarly high accuracy (ACC of 0.87) when using BERT and ALBERT, while ALBERT performs even better in terms of F1 score (0.76) and AUC (0.87). RoBERTa performs a little bit lower (ACC of 0.85), but the AUC is the same as BERT. From here, we can find that the improvement on text may have some degree of influence on the classification performance, such as increasing the parameters of BERT model and model size, etc., and the lightweight BERT model will reduce the classification accuracy. In addition, all image models have lower accuracy (ACC) and F1 scores (both 0.56 and 0.42) and lower AUC values (0.64-0.66). The performance of unimodal text classification is demonstrated to be significantly better than unimodal image classification, suggesting that discharge diagnosis information may be more valuable than NCCT image information in medical stroke prediction, but the Swin Transformer still outperforms the other transformer structures in terms of image classification alone, which may be attributed to the fact that the window interaction approach discovers some details in the image to better perceive the lesion. details in the image to get a better sense of the lesions and therefore improve the accuracy. In addition, the ablation experiments using different architectures found that under the premise of extracting text images using the same BERT, the effect of using SwinT is the same as that of using ViT, which may be due to the fact that the information perceived by Swin by sliding the window is overwritten by the text information extracted by the BERT. Therefore, we can conclude that the information derived from image information is limited, and with the constrained arithmetic, improving textual information extraction, as well as fusing more modalities may be the focus of future work conducted. This paper is the self-attention mechanism used for fusion, whether other fusion modalities with other fusion structures affect the results of the model is inconclusive, and there is a need for further research on more effective fusion methods and structures.

In addition to component-based issues, there is still an issue that cannot be ignored, which is the issue between text information and cross-semantic alignment of stroke images. Currently, we still place our hopes on the linear layer or the transf module with an attention mechanism. It has been observed that when there are ambiguous matches in some semantics, the prediction effect will be significantly reduced. The reason is still unclear. We will conduct more precise fine-grained alignment and calibration during cross-modal semantic alignment.

## V. CONCLUSION

In this work, we propose Multitrans, a multimodal fusion framework based on transformer architecture and self-attention mechanism, and study various transformer architecture-based methods based on NCCT scans and discharge diagnosis reports of patients undergoing stroke treatment. Performance of network modules in predicting functional outcome of stroke treatment. The results show that the performance of single-modal text classification is significantly better than that of single-modal image classification, but the effect of combined multi-modality is better than any single modality. This shows that although the Transformer model only performs poorly on imaging data, when combined with clinical meta-diagnostic information, both can learn better supplementary information and make a good contribution to accurately predicting stroke treatment effects. In future work, we plan to study and add more modal information into the converter model in our multimodal framework, and intend to make further in-depth work on optimizing the end-to-end multimodal model.